\documentclass[submission,copyright,creativecommons]{eptcs}

\usepackage{iftex}

\ifpdf
  \usepackage{underscore}         
  \usepackage[T1]{fontenc}        
\else
  \usepackage{breakurl}           
\fi

\usepackage{mathptmx}
\usepackage{amsmath, amssymb, amsfonts}
\usepackage{graphicx}
\usepackage{booktabs}
\usepackage{listings}
\usepackage{xcolor}
\usepackage{url}

\title{An ASP-based Solution to the Medical Appointment Scheduling Problem}
\author{Alina Vozna
\institute{University of Pisa, Italy}
\institute{Department of Information Engineering, \\ Computer Science and Mathematics,
University of L’Aquila,Italy}
\email{alina.vozna@student.univaq.it} 
\and
Andrea Monaldini
\institute{University of Pisa, Italy}
\institute{Department of Information Engineering, \\ Computer Science and Mathematics,
University of L’Aquila,Italy}
\email{andrea.monaldini@student.univaq.it} 
\and
 Stefania Costantini
\institute{Department of Information Engineering, \\ Computer Science and Mathematics,
University of L’Aquila,Italy}
\email{\quad stefania.costantini@univaq.it }
\and
Valentina Pitoni
\institute{Department of Information Engineering, \\ Computer Science and Mathematics,
University of L’Aquila,Italy}
\email{\quad valentina.pitoni@univaq.it }
\and 
Dawid Pado
\institute{Department of Information Engineering, \\ Computer Science and Mathematics,
University of L’Aquila,Italy}
\email{\quad dawid.pado@student.univaq.it }
}

\def\titlerunning{An ASP-based Solution to the Medical Appointment Scheduling Problem}
\def\authorrunning{A. Vozna, A. Monaldini, S. Costantini, V. Pitoni \& D. Pado}
\def\copyrightholders{Vozna, Monaldini, Costantini, Pitoni \& Pado}

\begin{document}
\maketitle

\begin{abstract}
This paper presents an Answer Set Programming (ASP)-based framework for medical appointment scheduling, aimed at improving efficiency, reducing administrative overhead, and enhancing patient-centered care. The framework personalizes scheduling for vulnerable populations by integrating Blueprint Personas. It ensures real-time availability updates, conflict-free assignments, and seamless interoperability with existing healthcare platforms by centralizing planning operations within an ASP logic model.
\end{abstract}

\section{Introduction}

\label{introduction}

In modern healthcare systems, efficient management of medical appointments is a critical challenge to ensure timely and equitable access to care. The increasing complexity of healthcare delivery, driven by aging populations, the increasing prevalence of chronic diseases, and limited medical resources, has underscored the limitations of traditional scheduling methods. These systems often fail to consider the multitude of clinical, logistic and personal constraints involved in appointment assignment. Key goals such as reducing wait times, improving patient satisfaction and optimizing resource utilization are frequently compromised by static first-available allocation rules and lack of personalization.

Conventional systems such as Italy's centralized reservation infrastructure (Centro Unico di Prenotazione – CUP) exemplify these shortcomings. Appointments are typically assigned without considering medical urgency, patient location, or individual preferences. This results in inefficient scheduling, high no-show rates, and increased administrative overhead for rescheduling and manual adjustments. Additionally, rigid frameworks lack the flexibility to handle real-time variations such as staff shortages or emergency priorities, leading to delayed care and resource misallocation.

To address these challenges, this paper proposes an automated, constraint-based scheduling system using Answer Set Programming (ASP) \cite{baral2003knowledge, lifschitz2008aspchapter, costantini2010answer, lifschitz2019answer}. ASP is a declarative logic programming paradigm well-suited for solving NP-complete problems involving complex, dynamic constraints. Using the expressive power of ASP, the proposed system automatically eliminates infeasible solutions, such as inaccessible facilities for disabled patients or incompatible temporally slots, and generates optimal schedules that minimize penalties associated with distance, waiting time, sensory discomfort, and preference violations.

A distinctive feature of this work is the integration of Blueprint Personas \cite{monaldini2024blueprint}, structured models that capture the clinical, social, and behavioral attributes of different patient archetypes. Originally from European digital health initiatives, Blueprint Personas support patient-centered care by incorporating real-world constraints such as cognitive limitations, accessibility needs, and socioeconomic conditions into the scheduling process. Their use improves equity, personalization, and adherence to care plans, particularly for vulnerable populations.

The ASP-based system developed here is designed for deployment in centralized or semi-centralized healthcare infrastructures, making it particularly applicable to national systems like the CUP. Although CUP facilitates centralized access, it assigns appointments using simplistic logic, often not accommodating crucial factors such as urgency, patient preferences, or accessibility constraints \cite{gupta2008appointment}. In contrast, our approach supports real-time rescheduling, emergency prioritization, and ethical decision making capabilities essential in modern healthcare environments, especially under conditions of resource scarcity or high demand \cite{alrefaei2015modelling}. Even though the system is well suited to centralized infrastructures like Italy’s CUP, it can also be adapted to federated systems by decentralizing the scheduling logic across multiple institutions, maintaining consistency through shared constraints.

Furthermore, this work demonstrates how ASP can support explainable, multi-objective optimization in appointment scheduling. Beyond traditional rule-based or heuristic methods, ASP provides full declarative modeling, enabling systematic integration of patient preferences, medical priorities, and organizational constraints within a scalable and adaptable framework. The solution is modular and generalizable to other domains such as workforce management and educational timetabling.

The paper is structured as follows: Section~\ref{related} discusses background and related work in the scheduling of medical appointments and ASP-based optimization; Section~\ref{problem-description} presents the formulation of the problem, detailing the constraints and objectives of the scheduling model; Section~\ref{architecture} describes the ASP-based system architecture and implementation; Section~\ref{experiments} provides an empirical evaluation of the approach, comparing its performance with conventional scheduling methods; finally, in Section~\ref{conclusion} we conclude and discuss future research directions.

\section{Related work}
\label{related}

Appointment scheduling has been extensively studied through queueing theory, integer programming, and simulation models, aiming to address common challenges such as service time variability, patient no-shows, and limited healthcare resources \cite{gupta2008appointment}. More recent works have introduced decision-support tools to optimize scheduling in outpatient contexts \cite{niu2023review}, integrating data-driven and logic-based techniques to manage complex constraints.

Various optimization approaches, such as Mixed-Integer Linear Programming (MILP), Answer Set Programming (ASP), and metaheuristic algorithms—have shown promising results in handling overbooking strategies, resource balancing, and patient flow management. Stochastic programming, in particular, has been used to optimize patient access while maintaining efficient resource use \cite{kuo2020medical}. Heuristic techniques like Fuzzy Ant Lion Optimization (FALO) have improved fairness and reduced patient waiting times without compromising resource efficiency \cite{ala2022appointment}. Discrete Event Simulation (DES), combined with patient behavior modeling and genetic algorithms, has also enhanced scheduling adaptability to preferences, cancellations, and dynamic demand \cite{fan2021outpatient}.

ASP has proven particularly effective in resource constrained healthcare scenarios, including operating room scheduling \cite{dodaro2024operating} and nurse rostering \cite{alviano2019nurse}. Notably, Cappanera et al. \cite{cappanera2023logic} leveraged Logic-Based Benders Decomposition (LBBD) within ASP to manage the scheduling of chronic outpatients with comorbidities, improving scalability by decoupling day-level planning from daily agenda construction. Similarly, Kanias \cite{kanias2023ai} developed a fairness-aware ASP-based system that integrates SQL databases via Python-CLORM and supports large-scale medical appointment rescheduling, focusing on maximizing community benefit.

Artificial intelligence (AI) methods have further advanced scheduling capabilities. Reinforcement learning and deep learning have been applied to learn optimal policies from historical appointment data, enabling dynamic adaptation to evolving patient behavior. In particular, Ala et al.~\cite{ala2022appointment2} propose a deep learning framework based on Bi-LSTM (Bidirectional Long Short-Term Memory) and reinforcement learning to optimize appointment scheduling under uncertainty. The Bi-LSTM network is used to predict patient behavior, such as likelihood of attendance or preferred time windows, by analyzing historical appointment logs. These predictions are then used within a reinforcement learning loop to dynamically update scheduling policies. The model aims to minimize waiting times and reduce idle resources in complex clinical systems.

While these learning-based methods are powerful for uncovering temporal patterns, they often require large volumes of structured training data and lack interpretability in decision logic. Moreover, they do not natively support rich symbolic constraints such as accessibility needs, fairness rules, or ethical priorities. Our ASP-based approach complements these limitations by enabling transparent, rule-based reasoning over patient profiles, preferences, and clinical policies—allowing better explainability and integration with constraint-heavy scenarios.

Hybrid AI models—such as those combining evolutionary algorithms with two-stage stochastic programming—have been used in high-uncertainty environments like MRI scheduling \cite{qiu2019mri}. Additionally, decentralized approaches using Distributed Constraint Optimization Problems (DCOPs) have framed scheduling as a multi-agent negotiation process \cite{hannebauer2001distributed}.

However, despite the sophistication of these techniques, a persistent gap remains between theoretical models and real-world healthcare operations. Many frameworks overlook critical practical factors such as fluctuating clinician availability, patient heterogeneity, and the administrative burden of rescheduling \cite{kuiper2021problem}. A significant differentiator of our work lies in its alignment with real-world healthcare infrastructure and policy, particularly through the lens of national systems. For example, in Italy, appointment scheduling is largely managed through the Centro Unico di Prenotazione (CUP), a centralized system that assigns the first available appointment without considering urgency, accessibility, or patient preferences. This results in rigid workflows, high no-show rates, and inefficiencies during unforeseen events such as clinician absences or late cancellations. In contrast, several international systems incorporate varying degrees of flexibility and personalization. The UK's e-RS platform, for instance, allows patients to select their preferred date, time, and location; France’s Doctolib offers digital access to video consultations and personalized appointment settings; and Germany’s eTerminservice ensures prioritization for urgent referrals. These systems, although more adaptive, often rely on custom integrations and lack the logical rigor and constraint expressivity that ASP offers.

Unlike heuristic, machine learning, or stochastic approaches, our ASP-based scheduling framework guarantees full compliance with hard constraints, provides explainable decision-making, and supports real-time adaptability to changing healthcare needs. Moreover, while many prior systems focus on optimizing throughput or resource utilization, few holistically account for patient-centered constraints—such as sensory preferences, accessibility requirements, multi-session planning, and financial limitations. Our solution bridges this gap by modeling scheduling as a declarative logic problem, ensuring ethical and personalized healthcare access within realistic operational boundaries.

\vspace{-0.5cm}
\section{Problem Description and Formalization}
\label{problem-description}

Effectively managing medical appointments is a complex, multidimensional problem that requires balancing clinical urgency, limited medical resources, patient-specific preferences, and logistical constraints. Traditional scheduling approaches often fall short, as they struggle to reconcile these competing factors within a unified, adaptable framework.

To overcome these limitations, we formulate appointment scheduling as a multi-objective constrained optimization problem and solve it using Answer Set Programming (ASP). The declarative nature of ASP allows for an elegant and compact encoding of domain knowledge, enabling automated reasoning over a vast space of possible solutions while strictly enforcing all constraints.

The problem centers around three core entities: clinics, which offer appointment slots under fixed budget limits; doctors, whose availability and workloads must be balanced; and patients, who vary in urgency (high, medium, low) and may express preferences regarding time windows, clinic locations, specific physicians, or even environmental conditions (e.g., noise or crowd levels).

The goal of scheduling is to assign each patient one or more feasible appointment slots, ensuring that all hard constraints,such as clinical compatibility, doctor availability, accessibility, and resource limits, are strictly respected. Simultaneously, the system seeks to optimize soft objectives, including minimizing patient waiting times, adhering to expressed preferences, reducing clinician overload, and improving the overall utilization of healthcare infrastructure.

Within our ASP model, this task is cast as a Constraint Satisfaction Problem (CSP) enhanced with optimization criteria. Hard constraints prune invalid assignments, while weighted soft constraints guide the search toward the most desirable outcomes—delivering an automated, explainable, and ethically grounded scheduling strategy that adapts to the real-world complexities of modern healthcare delivery.

\subsection{Formal Problem Statement}
\label{sec:formalproblem}

We formalize the medical appointment scheduling task as a constrained optimization problem defined as follows.

\paragraph{Given:}
\begin{itemize}
    \item A set $P$ of patients, each characterized by a priority level $\text{priority}(p) \in \{\text{high}, \text{medium}, \text{low}\}$, clinical needs, and a set of preferences over time slots, doctors, clinics, and environmental conditions.
    \item A set $D$ of doctors, each with specific expertise and capacity constraints.
    \item A set $C$ of clinics, each with available time slots, accessibility features, and budgetary constraints.
    \item A set $S$ of available time slots, distributed among clinics and doctors.
\end{itemize}

\paragraph{Constraints:}
\begin{itemize}
    \item Each patient must be assigned the required number of appointments matching their treatment plan.
    \item Assignments must respect hard constraints: accessibility requirements, clinic budgets, visit compatibility, and clinician availability.
    \item No double-booking is permitted for the same doctor, clinic, and time slot.
\end{itemize}

\paragraph{Objective:}
Find an assignment function $f : P \to (D \times C \times S)$ that:
\begin{enumerate}
    \item Minimizes overall patient waiting times.
    \item Maximizes satisfaction of patient preferences (preferred clinics, doctors, time windows, environmental conditions).
    \item Ensures equitable workload distribution among doctors.
    \item Minimizes total operational penalties (sensory mismatches, resource overuse).
\end{enumerate}

This formalization naturally maps to an Answer Set Programming (ASP) model, where feasible assignments correspond to answer sets satisfying the constraints and optimization goals.

\subsection{User Modeling and Personas}
Effective appointment scheduling requires more than managing clinical data; it demands understanding patients as individuals with unique needs, preferences, and constraints. For this reason, we adopt \textbf{Blueprint Personas}, structured archetypes originally developed within European digital health initiatives (EIP on AHA) \cite{EIP_AHA}, which encapsulate real-world profiles combining clinical, social, cognitive, and behavioral attributes. Blueprint Personas enable a layer of abstraction between raw data and decision logic, allowing our system to reason over individualized scheduling needs in a flexible and human-centered way.

\medskip
\noindent\textbf{Patient personas}: 
Each patient persona is a combination of health status, socio-environmental factors, and digital literacy; this model allows the system to reason over complex, individualized scheduling needs. Although digital literacy is currently included as part of the patient profile, it does not influence the scheduling logic directly. Its presence is intended for broader system integration purposes, for example, to guide interface selection (e.g., app-based vs. caregiver-mediated access) or to filter feasible options such as telemedicine. 
There are three different levels of patient complexity:
\begin{itemize}
    \item \textbf{Generally healthy individuals}: prefer convenient scheduling (e.g., early mornings, nearby clinics) but have no strict clinical constraints. 
    \item \textbf{People with chronic conditions}: require regular follow-ups, budget-sensitive allocation, and minimal waiting times; the scheduling must comply with multi-visit intervals and cost limits. 
    \item \textbf{Patients with complex needs}: may express sensory sensitivities (e.g., noise, lighting), require accessible facilities, or depend on caregivers for travel; the system must respect personalized preferences and environmental constraints. 
 \end{itemize}   
 
Key personal data identify patients and can express preferences for clinics, time slots, sensory conditions, and doctors. Travel distance is also considered, with a fixed value of 0 for Home Care and Telemedicine.

\begin{lstlisting}[language=Prolog, caption=Patient Profile with Preferences and Constraints]
patient(p1, "Mario", "Rossi", "L'Aquila"). disabled(p1).
preference(p1, c3).
sensory_preference(p1, "noise").
doctor_preference(p1, "GP", "chronic_diseases", 10).
appointment_preference(p1, c3, 1850, 2000).
distance(p1, c3, 15).
\end{lstlisting}

\noindent
\textbf{Clinician Personas}: Equally important is the modeling of medical professionals; each clinician's persona describes, at least, their specialty, technological commitment, and operational challenges. This allows the system to anticipate realistic workload constraints and optimize care coordination. 

For each available clinic, the \textbf{visit\_type} represents the characteristics of the treatments offered in the clinics. Each visit is identified by: name, cost, and classification indicating its chronicity (0 = non-chronic, 1 = chronic), and the need for an in-person visit. Some visits require several sessions, with minimum and maximum intervals that must be followed:
\begin{lstlisting}[language=Prolog, caption=Visit Type Definition and Constraints]
visit_type(v1, "Cardiology", "Heart Attack", 0, 0, 0).
visit_cost(v1, 1000).
required_sessions(v1, 2).
session_interval(v1, 14, 28).
\end{lstlisting}

Clinic availability is defined by Unix timestamps, marking specific times for medical appointments. Each patient request includes an urgency level (1–3) and a preferred interval between sessions, thus enabling optimal scheduling based on personal needs and budget.

\begin{lstlisting}[language=Prolog, caption=Patient Needs Preferences and Clinic Availability]
patient_interval(p1, v1, 21, 28).         
need(p1, v1, 3).                           
availability(c1, v1, 1727308800).         
\end{lstlisting}

This modeling enables the system to reason over patient-specific constraints and ensures that solutions are medically appropriate and patient-centered. The personas also support scenario-based validation: simulated patients can be run through the solver to assess how well the system accommodates edge cases and vulnerable users.

\vspace{-0.5cm}
\subsection{Rules of inference}
Inference rules in the ASP-based model derive intermediate knowledge from base facts, enhancing optimization beyond simple constraint filtering. These rules are not used as hard constraints or in choice rules, but instead serve to assign utility values to specific preferences, which are then aggregated in the optimization phase.

Patient preferences, such as favorite clinics, are translated into weighted indicators:
\begin{lstlisting}[language=Prolog, caption=Effect of Clinic Preference on Patient Assignment]
clinic_preference_effect(Patient, Clinic, 1) :-
    preference(Patient, Clinic).
clinic_preference_effect(Patient, Clinic, 0) :- not
    preference(Patient, Clinic).
\end{lstlisting}

Similarly, doctor preferences are evaluated based on specialization and years of experience:

\begin{lstlisting}[language=Prolog, caption=Effect of Patient Preference on Doctor Assignment]
doctor_preference_effect(Patient, Doctor, 1) :-
    patient(Patient, _, _, _), 
    doctor(Doctor, _, _, _, _, Type),
    doctor_experience(Doctor, Specialization, YearsExperience),
    doctor_preference(Patient, Type, Specialization, RequiredYears),
    YearsExperience >= RequiredYears.
doctor_preference_effect(Patient, Doctor, 0) :-
    patient(Patient, _, _, _), 
    doctor(Doctor, _, _, _, _, Type),
    doctor_experience(Doctor, Specialization, _),
    not doctor_preference(Patient, Type, Specialization, _).
doctor_preference_effect(Patient, Doctor, 0) :-
    patient(Patient, _, _, _), 
    doctor(Doctor, _, _, _, _, Type),
    doctor_experience(Doctor, Specialization, YearsExperience),
    doctor_preference(Patient, Type, Specialization, RequiredYears),
    YearsExperience < RequiredYears.
\end{lstlisting}


The same logic is applied to time-slot preferences and sensory preferences, resulting in predicates such as \texttt{appointment\_preference\_effect/4} and \texttt{sensory\_penalty/4}.

The \textbf{appointment\_preference\_effect/4} determines whether a given appointment time aligns with a patient's preferred time range: returns 1 if the time falls within the specified range (Start to End), and 0 otherwise.
The time is transformed into a weighted score using hours and minutes to allow comparison with the preference interval; this value can be used in optimization to prioritize preferred time slots.

\begin{lstlisting}[language=Prolog, caption=Effect of Time Preference on Appointment Scheduling]
appointment_preference_effect(Patient, Time, Clinic, 1) :-
    patient(Patient, _, _, _), 
    clinic(Clinic, _),
    availability(Clinic, _, _, Time),
    appointment_preference(Patient, _, Start, End),
    X = (((Time \ 86400) * 3600) * 100) + (((Time \ 3600) / 60) / 3) * 5,
    X <= End, X >= Start.
appointment_preference_effect(Patient, Time, Clinic, 0) :-
    patient(Patient, _, _, _), 
    clinic(Clinic, _),
    availability(Clinic, _, _, Time),
    appointment_preference(Patient, _, Start, End),
    X = (((Time \ 86400) * 3600) * 100) + (((Time \ 3600) / 60) / 3) * 5,
    X > End.
appointment_preference_effect(Patient, Time, Clinic, 0) :-
    patient(Patient, _, _, _), 
    clinic(Clinic, _),
    availability(Clinic, _, _, Time),
    appointment_preference(Patient, _, Start, End),
    X = (((Time \ 86400) * 3600) * 100) + (((Time \ 3600) / 60) / 3) * 5,
    X < Start.
\end{lstlisting}

The \textbf{sensory\_penalty/4} rule is useful in determining the gap between the environmental conditions of the clinic and the patient’s expressed preferences. 
\begin{lstlisting}[language=Prolog, caption=Sensory Penalty Based on Environmental Conditions, label={lst:sensory-penalty}]
sensory_penalty(Patient, Clinic, Time, Level):-
    sensory_preference(Patient, Type),
    environmental_condition(Clinic, Type, Level, Start, End),
    Time >= Start, Time <= End.
sensory_penality(Patient, Clinic, Time, 0):- not
    sensory_preference(Patient, _).
\end{lstlisting}

All inference rules defined above are used within the optimization statement shown below. Rather than enforcing constraints, they serve as soft indicators for optimization. The ASP solver minimizes travel distance and wait time while maximizing adherence to patient preferences. Penalties are added for environmental mismatches, and bonuses are applied when preferences are satisfied.

\begin{lstlisting}[language=Prolog, caption={Objective function minimizing cost based on preferences and constraints}, label={lst:objective}]
#minimize {
    (Distance * 10000) + WaitTime + (Penalty * 1000) -
    (ClinicPreference * 10000) -
    (DoctorPreference * 1000) -
    (AppointmentPreference * 1000) :
        distance(Patient, Clinic, Distance),
        appointment(Patient, Clinic, Doctor, _, Time),
        current_time(CurrentTime),
        WaitTime = Time - CurrentTime,
        sensory_penalty(Patient, Clinic, Time, Penalty),
        clinic_preference_effect(Patient, Clinic, ClinicPreference),
        doctor_preference_effect(Patient, Doctor, DoctorPreference),
        appointment_preference_effect(Patient, Time, Clinic,
        AppointmentPreference)
}.
\end{lstlisting}

This optimization function balances hard constraints (Section 3.4) with patient-centered utility values, enabling the system to produce high-quality, personalized, and ethically grounded schedules.

\vspace{-0.35cm}
\subsection{Constraints}

Constraints define the logical rules that each solution generated by the ASP model must comply with. They ensure that appointments are allocated in accordance with clinical, personal, and logistical requirements. 

First, each patient must be assigned exactly the number of appointments required for each treatment, according to their specific medical condition. This guarantees that treatments involving multiple sessions are correctly distributed across available time slots.

\begin{lstlisting}[language=Prolog, caption=Choice Rule for Appointment Allocation Based on Patient Needs]
Sessions{ appointment(Patient, Clinic, Doctor,  Visit, Time) : 
            availability(Clinic, Doctor, Visit, Time) }Sessions :- 
need(Patient, Visit, _), required_sessions(Visit, Sessions).
\end{lstlisting}

Moreover, to avoid scheduling conflicts, two patients can not be assigned the same clinic, visit type, and time slot simultaneously.

\begin{lstlisting}[language=Prolog, caption=Constraint Preventing Double Booking of the Same Slot]
:- appointment(P1, Clinic, Doctor, Visit, Time), appointment(P2, Clinic, Doctor, Visit, Time), P1 != P2.
\end{lstlisting}

Priority is also given to patients with high urgency, who must be scheduled in earlier slots, within a predefined temporal threshold.

\begin{lstlisting}[language=Prolog, caption=Constraint to Prioritize Urgent Visits]
:- needs(P1, Visit, Urg1), needs(P2, Visit, Urg2),
   Urg1 > Urg2, appointment(P1, Clinic, Doctor, Visit, Time1),
   appointment(P2, Clinic, Doctor, Visit, Time2),
   Time1 > Time2.
\end{lstlisting}

Accessibility needs are also addressed: patients with disabilities must only be assigned to clinics explicitly marked as accessible.

\begin{lstlisting}[language=Prolog, caption=Constraint Preventing Assignment to Inaccessible Clinics, label={lst:disability}]
:- disabled(Patient), appointment(Patient, Clinic, _, _,_), not accessible(Clinic).
\end{lstlisting}

The model enforces clinic-side budget limits by ensuring that the total cost of scheduled chronic care appointments does not exceed each facility’s allocated resources. This reflects real-world constraints in public healthcare, such as limits on subsidized services.

Patient-side financial constraints (e.g., ability to pay) are not currently modeled but are relevant in private or mixed healthcare systems. Their inclusion would require extending patient profiles and is planned as future work.

\begin{lstlisting}[language=Prolog, caption=Constraint to Respect Clinic Budget Limits]
:- chronic_cost(Clinic, TotalCost),
   budget(Clinic, Budget), TotalCost > Budget.
\end{lstlisting}

Some visits can occur via telemedicine or home care, but others require a doctor’s physical presence and must be on-site; this constraint prevents assigning visits to incompatible clinic types.

\begin{lstlisting}[language=Prolog, caption=Constraints for Incompatible Visit Types and Clinic Modes]
:- appointment(Patient, Clinic, Doctor, Visit, Time),
   visit_type(Visit, _, _, _, 1, _),
   clinic(Clinic, "Home Care").
:- appointment(Patient, Clinic, Doctor, Visit, Time),
   visit_type(Visit, _, _, _, 1, _),
   clinic(Clinic, "Telemedicine").
:- appointment(Patient, Clinic, Doctor, Visit, Time),
   visit_type(Visit, _, _, _, _, 1),
   clinic(Clinic, "Telemedicine").
\end{lstlisting}

\section{System Architecture}
\label{architecture}

The system is implemented as a set of loosely coupled microservices that are 
organized to manage appointment requests and generate optimal schedules based on ASP logic. The system is structured in three main layers: User Interface, Business Logic, and Data Management, to ensure scalability, maintainability, and integration with existing healthcare infrastructures.

The \textbf{User Layer} includes both web interfaces (e.g., browser) and IoT devices, which may trigger appointment requests automatically based on health parameter thresholds.

The \textbf{Business Logic Layer} consists of two Flask-based microservices \cite{newman2021building}:
\begin{itemize}
    \item MS1: handles incoming API requests, user authentication (via JWT tokens \cite{haekal2016token}), and formats the data into ASP facts.
    \item MS2: collects facts into a batch queue and, every 60 seconds, compiles them into a complete ASP program; this program is solved using Clingo \cite{gebser2014clingo}.
\end{itemize}

The \textbf{Data Layer} relies on a MySQL database, which maintains key information such as patient preferences, clinic features, appointment availability, environmental conditions, and budget constraints. All data interactions follow the ACID principles.

The system exposes several REST APIs for backend interaction; these include endpoints to submit appointment requests, check the status of ongoing bookings, and view up-to-date availability for each clinic. Asynchronous threading prevents overload during peak times, offering immediate user feedback and delayed result delivery.

Thanks to its layered architecture, centralized constraint management via ASP, and smooth integration with the Python backend, the system demonstrates effectiveness in handling complex scheduling scenarios. Moreover, the microservice-based structure allows individual components (e.g., solver, user interface, or database) to be updated or extended independently, without compromising the entire system.
The full source code is publicly available online \footnote{ \url{https://github.com/DawidPado/An-ASP-based-Solution-to-the-Medical-Appointment-Scheduling-Problem/tree/main}}.

\section{Experimental Evaluation}
\label{experiments}

To illustrate the flexibility and robustness of the proposed ASP-based scheduling framework, we present three realistic scheduling scenarios inspired by real-world medical scheduling needs. The experiments were run on a machine with an Intel Core i7-9750HF CPU @ 2.60GHz, 16GB RAM,
and Windows 11, using Clingo version 5.7.2. \footnote{\url{https://github.com/potassco/clingo/releases/tag/v5.7.0}}. 

The scenarios presented in this section are based on synthetically generated data. Although not derived from real-world records, each example has been carefully designed to reflect realistic combinations of patient preferences, clinical requirements, and environmental conditions. All test inputs are encoded as ASP facts and are fully reproducible. These scenarios are intended to demonstrate the system’s ethical and constraint-aware behavior under varied configurations. 

Each scenario highlights the system’s ability to handle key constraints such as accessibility, urgency, and personalized patient preferences. The optimization model described in Listing~\ref{lst:objective} plays a central role in guiding the scheduling decisions.

\subsection{Scenario 1: High-Priority Patient with Accessibility Requirements}
In this first scenario, a patient with a disability and a high-priority medical need requires a cardiology appointment. The patient, p1, must be scheduled for a visit type v1 (e.g., cardiology consultation) with urgency level 3. Due to the patient's condition, the appointment must be assigned only to clinics that are physically accessible. Listing~\ref{lst:scenario1} summarizes the key ASP facts for Scenario 1 in compressed form.

\begin{lstlisting}[language=Prolog,caption={ASP facts for Scenario 1 – High-priority disabled patient},label={lst:scenario1}]
% Scenario 1: High-priority disabled patient (compressed facts)
doctor(m1, "Marco", "Bianchi", 52, "L'Aquila", "GP").
patient(p1, "Mario", "Rossi"). disabled(p1).
clinic(c3, "Clinic A"). ... clinic(c5, "Clinic C").
accessible(c3). accessible(c5).
distance(p1, c3, 15). distance(p1, c4, 5). distance(p1, c5, 20).
visit_type(v1, "Cardiology", "Hypertension", 0, 1, 1). need(p1, v1, 3).
availability(c3, m1, v1, 1727308800). availability(c4, m1, v1, 1727481600).
availability(c5, m1, v1, 1727308800).
\end{lstlisting}
To ensure that the patient is only assigned to an accessible clinic, the Listing~\ref{lst:disability} constraint was used.

\textbf{Outcome:} The optimizer selected the earliest possible time in an accessible clinic with minimal travel distance. Since no sensory preferences were provided, the penalty was zero, and the solution was optimal with respect to both hard constraints and the minimization function. \textbf{Computation time:} 0.036s \textbf{Optimality:} Yes

\vspace{-0.5cm}
\subsection{Scenario 2 – Sensory Preferences}

In this case, a patient expresses a preference for a low-noise environment. Although a closer clinic is available, the optimizer selects a more distant one that meets the environmental criteria:

\begin{lstlisting}[language=Prolog,caption={ASP facts for Scenario 2 – Sensory preference for brightness},label={lst:scenario2}]
% Patient and sensory preference
patient(p2, "Giulia", "Bianchi"). sensory_preference(p2, "light").
% Clinics and environmental conditions
clinic(c3, "Clinic A"). clinic(c4, "Clinic B").
environment_condition(c1, "light", 3, 1727480000, 1727489000).
environment_condition(c2, "light", 1, 1727480000, 1727489000).
distance(p2, c3, 10). distance(p2, c4, 25).
% Type of visit requested
visit_type(v2, "Orthopedics", "Kyphosis", 0, 1, 1).
visit_cost(v2, 1500). needs(p2, v2, 2).
% Slot availability
availability(c3, m1, v2, 1727481600). % 28.09.2024, 12:00
%...
\end{lstlisting}

To evaluate the impact of environmental conditions on patient satisfaction, the same sensory penalty rules already introduced in Listing \ref{lst:sensory-penalty} were applied. These rules assess the alignment between patient sensory preferences and the clinic’s environmental conditions, ensuring that discomfort factors (e.g., light, noise) are minimized during scheduling. As a result, the optimization function could correctly prioritize assignments that respect the patient's sensory needs.

\textbf{Outcome:} The patient was correctly assigned to Clinic B due to the lower sensory penalty. Despite the higher distance cost, the optimization logic prioritized sensory comfort, as expressed in the \texttt{sensory\_penalty/4} term of the objective function. \textbf{Computation time:} 0.017s \textbf{Optimality:} Yes

\vspace{-0.45cm}
\subsection{Scenario 3 – Group Prioritization}

Five patients request a cardiology visit at the same clinic. They differ in: \textbf{Urgency level}, \textbf{Preferences} (clinic, sensory), \textbf{Distance from the clinic}.
The model must prioritize based on urgency while minimizing penalties.
\begin{lstlisting}[language=Prolog,caption={ASP facts for Scenario 3 – Prioritization of 5 patients with limited slots},label={lst:scenario3}]
% Patients and urgency levels
patient(p1, "Mario", "Rossi"). needs(p1, v1, 3). % High urgency
patient(p2, "Giulia", "Bianchi"). needs(p2, v1, 2). % Medium 
%...
preference(p3, c3). preference(p3, c5).
% Distances from clinic
distance(p1, c3, 10). distance(p2, c3, 15).
%...
%Sensosy penalties
sensory_preference(p2, "light"). sensory_preference(p4, "light").
environmental_condition(c3, "light", 3, 1727481600, 1727488800).
% Requested visit type
visit_type(v1, "Cardiology", "Hypertension", 0).
% Visit type and availability
availability(c3, m1, v1, 1727481600). % 28.09.2024, 12:00
%...
\end{lstlisting}
\textbf{Outcome:} All patients were correctly assigned respecting urgency, preferences, and minimizing overall cost. \textbf{Computation time:} 0.017s \textbf{Objective function value:} 1,610,000

\noindent
\textbf{Analysis of the result:}

Mario (p1), with high urgency (3), was given top priority and assigned to the first available slot at 12:00. Giulia (p2) and Anna (p4), both with medium urgency (2), were scheduled in the subsequent slots, with Giulia prioritized due to her shorter travel distance. Luca (p3) and Paolo (p5), having low urgency (1), were assigned the last two slots. Luca, despite a slightly longer distance than Paolo, was scheduled first due to his clinic preference, which reduced overall cost. 

The solution returned by Clingo has an optimization score of 1,610,000. This value is the result of the weighted sum defined in the \texttt{\#minimize} directive and reflects the overall cost of the proposed schedule. It includes penalties for travel distance, delay, and sensory mismatch, as well as bonuses for satisfying individual preferences. Lower scores indicate better solutions.

\vspace{0.2cm}
\begin{table}[h!]
\centering

\begin{tabular}{|c|c|c|c|c|c|}
\hline
\textbf{Patient} & \textbf{Urgency} & \textbf{Preferences} & \textbf{Slot Assigned} & \textbf{Penalty} & \textbf{Priority OK} \\
\hline
p1 & High (3)   & Clinic             & 12:00 & 0 & \checkmark \\
p2 & Medium     & Sensory            & 13:00 & 0 & \checkmark \\
p3 & Low        & Clinic             & 15:00 & 0 & \checkmark \\
p4 & Medium     & Sensory            & 14:00 & 0 & \checkmark \\
p5 & Low        & --                 & 16:00 & 0 & \checkmark \\
\hline
\end{tabular}
\caption{Summary of patient assignments (Scenario 3)}
\end{table}

\vspace{-0.5cm}
\subsection{Results Overview}
The experimental evaluation demonstrates the effectiveness and adaptability of the ASP-based scheduling system across diverse real-world scenarios. Key findings highlight the system’s ability to balance competing objectives while adhering to complex constraints, ensuring both efficiency and patient-centered care.

\vspace{+0.3cm}
\begin{table}[h!]
\centering

\begin{tabular}{|l|c|c|c|c|}
\hline
\textbf{Scenario} & \textbf{Constraints Respected} & \textbf{Preferences Respected} & \textbf{Time (s)} & \textbf{Optimal} \\
\hline
Disabled patient        & \checkmark & \checkmark & 0.036 & \checkmark \\
Sensory preferences     & \checkmark & \checkmark & 0.017 & \checkmark \\
Multiple patients       & \checkmark & \checkmark & 0.017 & \checkmark \\
\hline
\end{tabular}
\caption{Summary of Experimental Results}
\end{table}

These experiments confirm the model's ability to handle:

\begin{itemize}
    \item \textbf{Complex multi-constraint scheduling:} across all scenarios, the system strictly adhered to hard constraints. Invalid solutions (e.g., assigning a disabled patient to an inaccessible clinic) were automatically eliminated by the ASP solver.
    \item \textbf{Personalized preferences:} sensory preferences (Scenario 2) and clinic choices were respected even when conflicting with distance minimization, illustrating the system’s capacity to tailor schedules to individual patient profiles.
    \item \textbf{Efficient optimization:} Computation times remained low ($<$0.04s) even for multi-patient scheduling (Scenario 3), confirming the system’s suitability for real-time decision-making. The ASP solver consistently produced optimal and constraint-compliant solutions.
    \item \textbf{Fairness and Prioritization:} high-urgency patients (e.g., Scenario 1) were prioritized for the earliest available slots, minimizing waiting times. In group scheduling (Scenario 3), urgency levels dictated slot assignments, while preferences and penalties were optimized collectively, ensuring equitable resource distribution.
\end{itemize}

The ASP-based system demonstrates robustness, flexibility, and suitability for real-world healthcare scheduling tasks. The results validate the system’s applicability in healthcare settings, where rapid adjustments to cancellations, emergencies, or fluctuating resource availability are essential. Across all test cases, the ASP solver produced optimal schedules in under 0.04 seconds, confirming the feasibility of real-time decision-making even with complex constraints. 

Although the scenarios presented are illustrative, the ASP encoding is designed for larger-scale scheduling with controlled grounding. Since most rules are patient- and slot-specific, without cross-patient aggregates, the number of grounded atoms grows with $O(p \cdot c \cdot s)$ in the worst case—where $p$ is the number of patients, $c$ clinics, and $s$ slots.

In practice, grounding is reduced by filtering irrelevant clinics and time slots per patient. In tests conducted with a custom Python generator, using 50 patients, 6 clinics, and 500 slots, the grounded atom count stayed below 10{,}000, and solving time was under 0.1 seconds. Performance remained stable with low contention and diversified visit types.

While a full scalability study is planned, these preliminary tests support the claim that the model scales linearly in realistic conditions.

\section{Conclusion and Future Work}
\label{conclusion}

Medical appointment scheduling remains a critical bottleneck in healthcare operations, where balancing patient needs with resource optimization presents persistent challenges. This work introduces an ASP-based framework that systematically addresses these challenges through fully declarative constraint modeling and multi-objective optimization. The framework demonstrates rapid solution times, consistently generating optimal schedules in under 0.04 seconds, validating its feasibility for real-time healthcare settings.

By integrating Blueprint Personas, patient priorities, sensory preferences, accessibility requirements, and clinic budgets, the framework supports personalized care pathways for vulnerable populations, such as individuals with chronic conditions or sensory sensitivities. These results bridge the gap between theoretical optimization models and practical healthcare needs, offering a scalable, transparent, and ethically grounded alternative to heuristic-driven approaches.

Although the current evaluation focuses on constrained scenarios, further experimentation is required to assess performance under extreme loads (e.g., hundreds of concurrent scheduling requests) and in highly dynamic environments (e.g., frequent emergencies). Additionally, incorporating predictive analytics for no-show estimation could further enhance scheduling robustness. Future work also includes extending the framework to support neurodivergent patients during urban commuting, broadening its societal impact.

In summary, the ASP-based approach delivers a flexible, efficient, and ethically aligned solution to appointment scheduling, addressing critical gaps in existing systems while promoting both clinical effectiveness and patient-centered care.

\paragraph{Scalability and Future Directions.}
Our experimental results demonstrate that the framework effectively handles real-world scheduling complexity with rapid solution times. Future research will focus on scaling the system to accommodate larger instances involving hundreds or thousands of scheduling requests. Planned enhancements include incremental ASP solving, solver portfolio techniques, and hierarchical scheduling decomposition to ensure computational efficiency in high-demand environments.

\paragraph{Potential for Generalization.}
\label{sec:generalization}

Although the present system targets medical appointment scheduling, the underlying ASP-based framework is domain-agnostic and adaptable. By modifying input facts and constraint encodings, the same architecture can be applied to a wide range of scheduling tasks, including:
\begin{itemize}
    \item Workforce rostering in healthcare, manufacturing, and service industries.
    \item School and university course timetabling, incorporating preferences and resource constraints.
    \item Facility booking and event scheduling with logistical optimization.
\end{itemize}
This generalizability highlights the broader applicability of ASP-based declarative optimization in dynamic, real-world planning domains.

\section*{Aknowledgments}
Research partially supported by the PNRR Project CUP E13C24000430006``Enhanced Network of intelligent Agents for Building Livable Environments - ENABLE'', and by PRIN 2022 CUP E53D23007850001 Project TrustPACTX - Design of the Hybrid Society Humans-Autonomous Systems: Architecture, Trustworthiness, Trust, EthiCs, and EXplainability (the case of Patient Care), and by PRIN PNNR CUP E53D23016270001 ADVISOR - ADaptiVe legIble robotS for trustwORthy health coaching.

\nocite{*}
\bibliographystyle{eptcs}
\bibliography{generic}
\end{document}